\documentclass[letterpaper, 10 pt, conference]{ieeeconf}  

\IEEEoverridecommandlockouts                              
\usepackage{mathptmx} 
\usepackage{times}    
\usepackage{amsmath}  
\usepackage{amssymb}  
\usepackage{amstext}
\usepackage{graphicx}
\usepackage{multicol}
\usepackage[bookmarks=true]{hyperref}
\usepackage[usenames]{color}
\usepackage[x11names]{xcolor}
\usepackage{xspace}
\usepackage{balance}
\usepackage[lofdepth,lotdepth]{subfig}
\usepackage{pifont}

\makeatletter
\let\NAT@parse\undefined
\makeatother

\usepackage[numbers]{natbib}
\usepackage{tikz}
\usepackage{ifthen}
\usepackage{multirow}
\usepackage{textcomp}
\usepackage[export]{adjustbox}
\usetikzlibrary{matrix,calc}
\usetikzlibrary{shapes,arrows,chains}
\usetikzlibrary{positioning}
\usepackage[font={footnotesize}]{caption}
\usepackage{pgfplots}
\usepackage{makecell}

\usepgfplotslibrary{external}
\DeclareMathAlphabet{\mathcal}{OMS}{cmsy}{m}{n}

\usepackage{typed-checklist} 

\title{\LARGE \bf
Spatial Reasoning via Deep Vision Models for Robotic Sequential Manipulation
}

\author{%
  Hongyou Zhou \rlap{\textsuperscript{1}}
  \and
  Ingmar Schubert \rlap{\textsuperscript{1}}%
  \and
  Marc Toussaint \rlap{\textsuperscript{1,2}}%
  \and
  Ozgur S. Oguz \rlap{\textsuperscript{3}} 
}

\begin{document}

\maketitle
\thispagestyle{empty}
\pagestyle{empty}

\footnotetext[1]{Technical University of Berlin}
\footnotetext[2]{Science of Intelligence Cluster of Excellence, TUB}
\footnotetext[3]{Computer Eng. Dept., Bilkent University. This work was supported by TUBITAK under 2232 program with project number 121C148 (``LiRA``).}

\begin{abstract} 
In this paper, we propose using deep neural architectures (i.e., vision transformers and ResNet) as heuristics for sequential decision-making in robotic manipulation problems. 
This formulation enables predicting the subset of objects that are relevant for completing a task.
Such problems are often addressed by task and motion planning (TAMP) formulations combining symbolic reasoning and continuous motion planning.
In essence, the action-object relationships are resolved for discrete, symbolic decisions that are used to solve manipulation motions (e.g., via nonlinear trajectory optimization).
However, solving long-horizon tasks requires consideration of all possible action-object combinations which limits the scalability of TAMP approaches. 
To overcome this combinatorial complexity, we introduce a visual perception module integrated with a TAMP-solver.
Given a task and an initial image of the scene, the learned model outputs the relevancy of objects to accomplish the task. 
By incorporating the predictions of the model into a TAMP formulation as a heuristic, the size of the search space is significantly reduced.
Results show that our framework finds feasible solutions more efficiently when compared to a state-of-the-art TAMP solver. 
\end{abstract}

\tikzstyle{arrow} = [draw, -latex']
\tikzstyle{line} = [draw, -]
\tikzstyle{input} = [rectangle, draw, fill=white!20, 
    text width=5em, text centered, rounded corners, minimum height=1cm, minimum width=2cm]
\tikzstyle{cnn} = [rectangle, draw, fill=white!20, 
    text width=1em, text centered, rounded corners, minimum height=2.5cm, minimum width=0.5cm]
\tikzstyle{resnetl} = [rectangle, draw, fill=white!20, 
    text width=1em, text centered, rounded corners, minimum height=2cm, minimum width=0.5cm]
\tikzstyle{resnets} = [rectangle, draw, fill=white!20, 
    text width=1em, text centered, rounded corners, minimum height=1cm, minimum width=0.5cm]
\tikzstyle{t} = [rectangle, draw, fill=white!20, 
    text width=1em, text centered, rounded corners, minimum height=.5cm, minimum width=.5cm]
\tikzstyle{output_sigma} = [circle, draw, fill=white!20, 
    text width=1em, text centered, rounded corners, minimum size=.2cm]

\section{Introduction}
Task and Motion Planning (TAMP) usually combines a discrete search on a symbolic, logical level with a geometric planner. 
Discrete search is employed to find the high-level symbolic action sequence, which, in turn, informs the geometric planner to solve for a feasible motion path that satisfies the goal state.
Due to the huge computational overhead of solving geometric problems and the combinatorial complexity of discrete decisions, solving a TAMP problem often requires a large amount of computation~\cite{wells2019learning,toussaint2018differentiable,rodriguez2019iteratively,driess2019hierarchical}.

The process of task planning often depends on the number of objects that are available in the environment and the operations that the robot can perform. 
The algorithm builds the corresponding search tree based on the combination of objects and robot action set and expands the tree step by step (with action-object(s) tuples) until it reaches the goal state. 
Exponential growth rate of the search tree tremendously diminishes the efficiency of the TAMP methods. 
In reality, there are often multiple objects in the same environment, and a large number of operational possibilities by the autonomous agent, which impede the usability of TAMP solvers in such complex scenarios.
Fortunately though, the task can usually be achieved by only considering a small subset of these objects.

Machine learning methods using visual data offer efficient ways to solve robotic manipulation problems~\cite{kase2020transferable, driess2021learning}.
However, one challenge in integrating machine learning into TAMP is how to set up the problem (for example) by encoding objects in the scene as fixed-length inputs and then feeding them to the learning algorithm~\cite{driess2020deep}. 
In general, learning methods are often integrated into robotic problems for a single task only, where fixed-size feature representation is often sufficient. 
On the other hand, TAMP formulations are designed to tackle a broad set of tasks in different environments with varying numbers of objects and goals.
This hinders the usage of such fixed size features for solving sequential robotic manipulation problems.

In this study, we argue that task-relevant objects can be inferred directly from the sensor space of the robot, instead of in a feature space. 
The sensor space has a fixed dimensionality, making it suitable for standard learning algorithms. Furthermore, the sensory information provides details about what the robot can extract from the current world state. 
However, it does not account for historical data or other prior information.
Therefore, this paper aims to answer the following question: Is it is possible to learn the relevance of an object in the scene given the goal state? 
This is critical because, as mentioned previously, the complexity of the TAMP algorithm belongs to the NP-hard class, and therefore the efficiency of the algorithm can be greatly improved if its search space can be reduced by learning methods.

\begin{figure} 
\centering
\begin{minipage}[c]{.52\linewidth}
\centering
\subfloat[Scene]{\includegraphics[width=\linewidth]{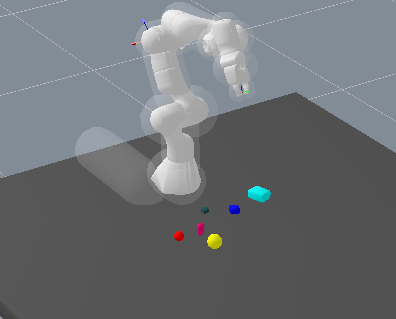}}
\end{minipage}%
\ 
\begin{minipage}[c]{.45\linewidth}
\centering
\subfloat[Scene from camera]{\includegraphics[width=.8\linewidth]{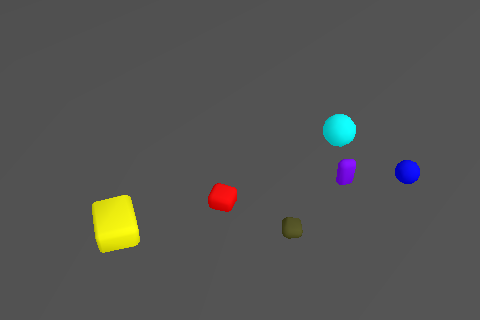}} \\
\subfloat[]{\includegraphics[width=.15\linewidth]{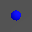}} \hspace{1mm}%
\subfloat[]{\includegraphics[width=.15\linewidth]{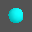}} \hspace{1mm}%
\subfloat[]{\includegraphics[width=.15\linewidth]{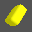}}
\end{minipage}
\caption{Test scenario with six objects: (a) shows the overview of this test scenario, (b) is the view from the camera located on the robot end-effector, (c, d and e) are the canonical views of objects, the first two are related to the goal predicate, and the third is the object whose relevance is queried. Take the predicate \textit{(on-left-side)} as an example, using these canonical views as inputs would query the model whether object in (e) should be considered when the goal is to \textit{put object in (c) on the left-side of object in (d)}.
}
\label{fig:1}
\end{figure}

We propose a deep neural network that takes a scene image, canonical views of objects included in the goal predicate, and the canonical view of another object in the scene to predict the relevance of that object for achieving the task (Fig.~\ref{fig:1}). 
The relevance of objects can then be used to improve the time complexity of mixed-integer programs, which is one way to realize TAMP, where the discrete action (integer) variable represents the abstract decision and the resulting nonlinear trajectory optimization problem represents the geometric part. 
In the experiments, we consider the problem of grasping the target object with one robot arm and placing it on the position that is indicated by the goal predicate, as shown in Fig.~\ref{fig:1}. 
An object can be considered as irrelevant due the non-interference of the trajectory of reaching the goal state. In contrast, if an object blocks the trajectory (i.e., there might be a collision), then the symbolic planner has to offer a different solution that instructs the robot to move this object out of the current place, which in turn makes this object relevant.

One of the experiment is shown in Fig.~\ref{fig:1}. 
As an example, if our goal is to place object \textit{c} to the left of object \textit{d}, then the symbolic planner, without prior knowledge, will traverse all possibilities and eventually find a path that satisfies the ``object \textit{c} is to the left of object \textit{d}'' state, or if there is no possible solution, then it returns non-reachable. 
However, as humans, we can quickly evaluate that no other object needs to be moved except for these two objects.
This is what we want to achieve with our trained classifier, thus greatly limiting the expansion of the search space.
In general, completing the goal requires actions on very few objects, and the majority of the objects that appear in the scene are irrelevant to the goal state. 
Hence, we train several neural network architectures, including vision transformers (ViT) and residual neural networks (ResNet), to predict the relevancy of objects.
The learned classifier then guides the discrete search. 
The main contributions are as follows:
\begin{itemize}
    \item We propose using vision-based deep neural networks to predict the relevance of the objects from visual input for sequential manipulation planning.
    \item We propose a flexible and scalable \textit{predicate extractor} to represent the spatial relationship.
    \item We develop the use of the network to guide the discrete search as a heuristic.
\end{itemize}
We support those contributions by showing that the trained models reach high accuracy, and our approach significantly reduces the number of motion planning problems in general, and thus improving computational efficiency of a state-of-the-art TAMP solver.
By evaluating multiple deep neural network models on the same dataset, we did not observe significant performance differences, e.g., using a ViT does not appear to be a better choice~\cite{yuan2022sornet}.

\tikzstyle{context_patches} = [rectangle, draw, fill=blue!20, inner sep=0.0mm,
    text width=10em, text height = 1.8cm, text centered, rounded corners, minimum height=2cm, minimum width=11cm]
\tikzstyle{context_image} = [rectangle, draw, fill=orange!20, inner sep=0.0mm,
    text width=20em, text centered, rounded corners=2pt, minimum height=1.2cm, minimum width=10cm]
\tikzstyle{context_patch} = [rectangle, draw, fill=orange!20, inner sep=0.0mm,
    text width=0em, text centered, rounded corners=2pt, minimum height=0.5cm, minimum width=0.6cm]
\tikzstyle{canonical_views} = [rectangle, draw, fill=red!20, inner sep=0.0mm,
    text width=10em, text height = 1.8cm, text centered, rounded corners, minimum height=2cm, minimum width=5.1cm]
\tikzstyle{goal_pred} = [rectangle, draw, fill=gray!15, inner sep=0.0mm,
    text width=5em, text height = 1cm, text centered, rounded corners, minimum height=1.2cm, minimum width=2.3cm]
\tikzstyle{query} = [rectangle, draw, fill=gray!5, inner sep=0.0mm,
    text width=4em, text height = 1cm, text centered, rounded corners, minimum height=1.2cm, minimum width=0.8cm]
\tikzstyle{resnet_large} = [rectangle, draw, fill=orange!20, inner sep=0.0mm,
    text width=20em, text centered, rounded corners, minimum height=1cm, minimum width=11cm]
\tikzstyle{resnet_small} = [rectangle, draw, fill=orange!20, inner sep=0.0mm,
    text width=10em, text centered, rounded corners, minimum height=1cm, minimum width=5.1cm]
\tikzstyle{projection_patches} = [rectangle, draw, fill=orange!20, inner sep=0.0mm,
    text width=20em, text centered, rounded corners, minimum height=1cm, minimum width=16.5cm]
\tikzstyle{projection_patch} = [rectangle, draw, fill=orange!20, inner sep=0.0mm,
    text width=0em, text centered, rounded corners=2pt, minimum height=0.5cm, minimum width=0.3cm]
\tikzstyle{projection_patch_description} = [rectangle, draw, fill=red!20, inner sep=0.0mm,
    text width=0.2cm, text centered, rounded corners=2pt, minimum height=0.5cm, minimum width=0.3cm]
\tikzstyle{transformer} = [rectangle, draw, fill=gray!20, inner sep=0.0mm,
    text width=20em, text centered, rounded corners, minimum height=1cm, minimum width=16.5cm]
\tikzstyle{embedding} = [rectangle, draw, fill=purple!20, inner sep=0.0mm,
    text width=0.2cm, text centered, rounded corners=2pt, minimum height=0.809cm, minimum width=0.3cm]
\tikzstyle{readout} = [rectangle, draw, fill=green!10, dashed,
    text width=20em, text depth = 2.6cm, anchor=north, text centered, rounded corners, minimum height=2.6cm, minimum width=16.5cm]
\tikzstyle{dataset} = [rectangle, draw, fill=lime!10, text=black, dashed,
    text width=20em, text depth = 2.6cm, anchor=north, text centered, rounded corners, minimum height=2cm, minimum width=16.5cm]
\tikzstyle{mlp} = [rectangle, draw, fill=white!20, 
    text width=5em, text centered, rounded corners, minimum height=1cm, minimum width=3.4cm]
\tikzstyle{output} = [rectangle, draw, fill=lime!20, 
    text width=5em, text centered, rounded corners, minimum height=1cm, minimum width=2cm]
\tikzstyle{arrow_txt} = [text width=2cm,midway,right=0em,align=left, font={\fontsize{8pt}{12}\selectfont}]

\section{Related work}
\subsection{Task and Motion Planning}
A TAMP problem is typically specified in two parts: the first part is a symbolic discrete decision, and the second part is a continuous motion planner. 
The symbolic discrete decision proposes a discrete node to reach the goal state, while the continuous motion planner is intended to solve the feasibility problem between discrete states. 
There are three mainstream solutions for a motion planner: sampling based~\cite{alili2010interleaving, srivastava2014combined, de2013towards, dantam2018incremental, simeon2004manipulation}, discretization based on the configuration/action space~\cite{erdem2011combining, lagriffoul2012constraint, lagriffoul2014efficiently, lozano2014constraint}, and the optimization based methods~\cite{shoukry2016scalable, hadfield2016sequential}. 
Logic Geometric Programming (LGP)~\cite{toussaint2015logic} also falls into the optimization-based category, where the logic imposes a skeleton of active constraints on a nonlinear program~\cite{toussaint2020describing, ha2020probabilistic}. 
Formalizing this hybrid nature of the contact and interaction modes in robotics  results in a mixed-integer program~\cite{hogan2018reactive, deits2014footstep}, and LGP is formulated as such for TAMP problems~\cite{toussaint2018differentiable}. 
A major problem of LGP, though, is the combinatorial complexity that is introduced by the discrete variables (i.e., due to the number of actions or objects in the scene). 
In this study, we propose to address this complexity by reducing the search space using a vision-based deep learning model that predicts a small set of objects to be interacted with to realize a task.


\subsection{Learning for Task and Motion Planning}
Advances in deep networks have made it possible to build a planning system that is based on sensor inputs. 
For example, attention mechanisms offer unique opportunities to learn representations of the environment~\cite{vaswani2017attention, dosovitskiy2020image, carion2020end, Guo20222ndPS}. 
Such models can be employed to reason about the spatial relationship between objects, thus the symbolic planner can leverage this information for high level planning~\cite{yuan2022sornet, nguyen2020self}.
Our model is based on vision transformer and residual neural network~\cite{vaswani2017attention, he2016deep}, and uses supervised learning to give the model the ability to infer task-relevant objects based on an image in the scene. 
The aim is to use visual inference to guide the symbolic planner to consider the option that is more likely to succeed in a given situation.
The viability of image-based deep networks for direct service to planning and control systems has also been confirmed by many studies~\cite{kase2020transferable, driess2020deep, driess2021learning}. 

In spirit, the motivation with SORNET~\cite{yuan2022sornet} is most similar to ours. 
The input to SORNET comprises image patches and canonical views of the objects to be queried. 
The output is the spatial relationship between the two objects queried in the scene.
However, this information is often not directly used in task and motion planning, because knowing the spatial relationship between objects does not help much in motion planning. 
SORNET uses the embedding output of the vision transformer (ViT) to train additional networks that models the predicates.
We, instead, use ViT to directly train a classifier that outputs the probability of relevancy of objects for a given goal.
This allows us to directly integrate the model output into our TAMP formulation.


In general, the method provided in this study inferred the task-related objects on the basis of images, which can limit the space of the discrete search part of task and motion planning, thus improving the overall search efficiency.
\section{MIXED-INTEGER PROGRAM FOR ROBOT TRAJECTORY PLANNING}
In this section, we explain our task and motion planning formulation based on ~\cite{toussaint2015logic, toussaint2017multi}.
Use cases for this formulation are two-fold: (\textit{i}) we solve a large number of complex rearrangement problems for obtaining labels of task-relevancy of objects,
(\textit{ii}) we integrate our learned classifiers for improving the computational efficiency of solving such sequential manipulation problems through the trajectory optimization framework that we utilize to create labels as training data and supervise deep learning as it solves the existing problem. 


Consider $s$ a symbolic state variable drawn from the set $S$. 
This set $S$ characterizes the scene, which is formulated by $\mathcal{X} \subset \mathbb{R}^{n(S)} \times SE(3)^{m(s, S)} \times \mathbb{R}^{6 \cdot n_{\textrm{cp}}(s, S)}$. 
Here, the scene configuration integrates robot joint positions ($n(S)$-dimensional), poses of $m(s, S)$ rigid objects, and the dynamics of wrench interactions at the contact points ($n_{\textrm{cp}}(s, S)$). 
To identify a globally consistent and feasible path $x: [0, KT] \rightarrow \mathcal{X}(s_{k(t)}, S)$ within this configuration, our objective is to minimize


\begin{subequations}\label{eq:1}
\begin{align}
P(g, S) = \min_{\substack{ K \in \mathbb{N} \\ x: [0, KT] \rightarrow \mathcal{X} \\ a_{1:K}, s_{1:K}}} & \int_{0}^{KT} c( x(t), \dot{x}(t), \ddot{x}(t), s_{k(t)}, S)dt \label{eq:1a} \\
        s.t. \qquad & \nonumber \\
        \forall_{t \in [0, KT]} : \ & h_{\textrm{eq}}(x(t), \dot{x}(t), s_{k(t)}, S) = 0 \label{eq:1b} \\
        \forall_{t \in [0, KT]} : \ & h_{\textrm{ineq}}(x(t), \dot{x}(t), s_{k(t)}, S) \leq 0 \label{eq:1c}\\
        \forall_{k=1,...,K} : \ & h_{\textrm{sw}}(x(kT), \dot{x}(kT), a_{k}, S) = 0 \label{eq:1d}\\
        \forall_{k=1,...,K} : \ & a_{k} \in \mathbb{A}(s_{k-1}, S) \label{eq:1e}\\
        \forall_{k=1,...,K} : \ & s_{k} = succ(s_{k-1}, a_{k}) \label{eq:1f}\\
                              & x(0) = \tilde{x}_{0}(S) \label{eq:1g}\\
                              & s_{0} = \tilde{s}_{0}(S) \label{eq:1h}\\
                              & s_{K} \in \mathcal{S}_{\textrm{goal}}(g). \label{eq:1i}
\end{align}
\end{subequations}

There are $K \in \mathbb{N}$ phases, each having a uniform duration $T > 0$.
We assume global continuity for the path $(x~\in~C([0,~TK]))$, and smoothness for each phase $x~\in~C^{2}([(k-1)T,~kT])$.
Note that the value of $K$ is part of the decision problem itself.
These phases are often referred to as kinematic modes, as mentioned in \cite{toussaint2018differentiable,mason1985mechanics}. 
Notably, as the number of degrees of freedom for the objects and the number of contact interactions vary in accordance with the symbolic state $s_{k(t)}$, where $k(t)=\left\lfloor{t/T}\right\rfloor$, the dimensionality of the path can also fluctuate between different stages.


Equations \eqref{eq:1a}, \eqref{eq:1b}, and \eqref{eq:1c} collectively define the waypoint in phase $k$ of the entire path. 
Here, $s_{k} \in \mathcal{S}(S)$ represents the state within the symbolic domain. 
Switching constraints between various kinematic modes are characterized by ~\eqref{eq:1d}. 
The end of each path phase is marked by an action transition ~\eqref{eq:1f}. 
It is important to note that we make the assumption that $c, h_{eq}$ and $h_{ineq}$ are differentiable for a fixed $s$.


The discrete action $a_{k}$~\eqref{eq:1e} paired with an object is called an action-object tuple. 
During tree search to find the action sequence, LGP expands the current node based on a possible action-object tuple. 
Once the scene becomes complex, the number of possible combinations of action-objects grows exponentially. 
This makes traversing the whole search space infeasible.
Adding reasonable prior knowledge then becomes critical to prune irrelevant nodes.




Given an initial symbolic state $s_{0}=\tilde{s}_{0}(S)$~\eqref{eq:1h}, the action sequence $a_{1:K}$ establishes the sequence of symbolic states $s_{0:K}$. The goal state is symbolically defined by the set $\mathcal{S}_{\textrm{goal}}(g)$, where the symbolic goal is represented by $g \in \mathbb{G}(S)$.

A solution to the LGP is represented by an action sequence, denoted as $a_{1:K}$, which guides the symbolic states from their initial configuration towards the symbolic goal state. Additionally, it is essential to ensure that the constraints within and between each stage are satisfied.

We characterize the feasibility of an action sequence $a_{1:K}=(a_{1},...,a_{K})$ through the presence of a corresponding path,
\begin{equation} \label{eq:2}
F_{S}(a_{1:K}) =
    \begin{cases}
        1,  & \text{if } \exists x : [0, KT] \rightarrow \mathcal{X} : \eqref{eq:1b} - \eqref{eq:1h}\\
        0,  & \text{otherwise}
    \end{cases}.
\end{equation}
We also define a function that describes whether an object is involved in solving the task,
\begin{equation} \label{eq:3}
Z_{S}(o) =
    \begin{cases}
        1,  & \text{if } F_{S}(a_{1:K}) = 1 \rightarrow \exists a_{k \in K}: \theta(a_{k}) = o\\
        0,  & \text{otherwise}
    \end{cases}.
\end{equation}
$\theta(a)$ takes an action as parameter and retrieves the object involved in the action. All actions have only one operation object, thus, the outcome of $\theta$ is a single object.

The comprehensive LGP formulation, as given in \eqref{eq:1}, aims to identify not only a viable solution but also a sequence of actions that results in the lowest trajectory costs, as expressed in \eqref{eq:1a}, when compared to all alternative sequences leading to the goal.
A viable solution is referred to as a \textit{solution to the TAMP/LGP problem}, while a solution that minimizes the cost is termed an \textit{optimal solution of the LGP}.
Given the multitude of possible discrete action sequences, rather than the optimal solutions, our primary focus is on acquiring an optimal solution.


Equation ~\eqref{eq:2} represents the feasibility of the nonlinear program~\eqref{eq:1} mathematically. 
In practice, though, ~\eqref{eq:1} is solved by discretizing $x$ and using a Gauss-Newton based augmented-Lagrangian method.
Specifically, the action sequence $F_{S}(a_{1:K})$ is constructed such that violations $(h_{eq} \neq 0, h_{ineq} > 0)$ are allowed below a predefined threshold for each waypoint. 
If these violations exceed the threshold, the problem is deemed infeasible.

\begin{figure*}
\centering
\begin{minipage}[c]{.15\linewidth}
\centering
\begin{tikzpicture}[node distance = 0cm, auto]
\node (scene_img)[inner sep=0pt] {\includegraphics[width=\linewidth]{img/scene.png}};
\node (scene_img_caption) [above=0cm of scene_img] {\footnotesize scene $S$};
\end{tikzpicture}
\end{minipage}%
\ 
\begin{minipage}[c]{.8\linewidth}
\centering
\begin{tikzpicture}[node distance = 0cm, auto]
\node (scene_img) [input] {\includegraphics[width=0.9\linewidth]{img/scene_img.png}};
\node (scene_img_caption) [above=-0.1cm of scene_img] {\footnotesize input scene image};

\node (canonical_view) [input, below=0.5cm of scene_img, fill=red!20] 
    {\includegraphics[width=0.25\linewidth]{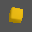}
     \includegraphics[width=0.25\linewidth]{img/canonical_obj3.png}
     \includegraphics[width=0.25\linewidth]{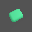}};
\node (canonical_view_caption) [above=-0.1cm of canonical_view] {\footnotesize canonical view};
     
\node (scene_img_patches) [input, right=0.5cm of scene_img, fill=blue!20] { 
    \adjincludegraphics[width=.25\linewidth,trim={{.0\width} {.66\height} {.66\width} {.0\height}},clip]{img/scene_img.png}
    \adjincludegraphics[width=.25\linewidth,trim={{.33\width} {.66\height} {.33\width} {.0\height}},clip]{img/scene_img.png}
    \adjincludegraphics[width=.25\linewidth,trim={{.66\width} {.66\height} {.0\width} {.0\height}},clip]{img/scene_img.png}
    
    \adjincludegraphics[width=.25\linewidth,trim={{.0\width} {.33\height} {.66\width} {.33\height}},clip]{img/scene_img.png}
    \adjincludegraphics[width=.25\linewidth,trim={{.33\width} {.33\height} {.33\width} {.33\height}},clip]{img/scene_img.png}
    \adjincludegraphics[width=.25\linewidth,trim={{.66\width} {.33\height} {.0\width} {.33\height}},clip]{img/scene_img.png}
    
    \adjincludegraphics[width=.25\linewidth,trim={{.0\width} {.0\height} {.66\width} {.66\height}},clip]{img/scene_img.png}
    \adjincludegraphics[width=.25\linewidth,trim={{.33\width} {.0\height} {.33\width} {.66\height}},clip]{img/scene_img.png}
    \adjincludegraphics[width=.25\linewidth,trim={{.66\width} {.0\height} {.0\width} {.66\height}},clip]{img/scene_img.png}
    };
\node (scene_img_patches_caption) [above=-0.1cm of scene_img_patches] {\footnotesize scene image patches};
    
\node (cnn) [cnn, right=0.5cm of scene_img_patches, yshift=-0.55cm, fill=orange!20] {\rotatebox{90}{\footnotesize CNN}};
\node (label) [cnn, right=0.5cm of cnn] {\rotatebox{90}{\footnotesize Attach position label}};
\node (transformer) [cnn, right=0.5cm of label, fill=gray!20] {\rotatebox{90}{\footnotesize Transformer}};
\node (t1) [t, right=0.5cm of transformer, yshift=0.6cm, fill=red!50] {\rotatebox{90}{}};
\node (t2) [t, right=0.5cm of transformer, yshift=0cm, fill=green!50] {\rotatebox{90}{}};
\node (t3) [t, right=0.5cm of transformer, yshift=-0.6cm, fill=cyan!50] {\rotatebox{90}{}};
\node (embedding_caption) [above=0.01cm of t1] {\scriptsize obj embedding};
\node (readout) [cnn, right=0.5cm of t2, yshift=0cm, fill=green!10] {\rotatebox{90}{\footnotesize Predicate Extractor}};
\node (output_sigma) [output_sigma, right=0.5cm of readout] {\footnotesize $\sigma$};
\node (output_sigma_caption) [right=0.1cm of output_sigma] {\footnotesize $t(I, p_{0}, p_{1}, q)$};

\path [arrow] (scene_img.east) -- (scene_img_patches.west);
\path [arrow] (scene_img_patches.east) -- (scene_img_patches-|cnn.west);
\path [arrow] (canonical_view.east) -- (canonical_view-|cnn.west);
\path [arrow] (cnn.east) -- (label.west);
\path [arrow] (label.east) -- (transformer.west);
\path [arrow] ([yshift=0.6cm,xshift=0cm]transformer.east) -- (t1.west);
\path [arrow] ([yshift=0cm,xshift=0cm]transformer.east) -- (t2.west);
\path [arrow] ([yshift=-0.6cm,xshift=0cm]transformer.east) -- (t3.west);

\path [arrow] (t1.east) -- ([yshift=0.6cm,xshift=0cm]readout.west);
\path [arrow] (t2.east) -- ([yshift=0cm,xshift=0cm]readout.west);
\path [arrow] (t3.east) -- ([yshift=-0.6cm,xshift=0cm]readout.west);

\path [arrow] (readout.east) -- (output_sigma.west);

\end{tikzpicture}
\end{minipage}
\caption{Proposed neural net architecture based on transformer to predict the involvement of objects in the physical interaction for a given goal predicate.
}
\label{fig:2}
\end{figure*}

\begin{figure}[!h]
\centering
\resizebox{1\linewidth}{!}{
\begin{tikzpicture}[node distance = 0cm, auto]
    \node (cps) [context_patches] {Context Patches};
    \node (cvs) [canonical_views, right=0.4cm of cps] {Canonical Object Views};
    \node (gp) [goal_pred, above=-1.28cm of cvs, xshift=-0.85cm] {Predicate};
    \node (q) [query, right=0.4cm of gp] {Query};
    \node (pps) [projection_patches, above=0.3cm of cps, xshift=2.75cm] {Linear Projection of Flattened Patches};
    
    \node (cp1) [context_patch, above=0cm of pps, xshift=-7.15cm, yshift=-2cm] {};
    \node (cp2) [context_patch, right=0.5cm of cp1] {};
    \node (cp3) [context_patch, right=0.5cm of cp2] {};
    \node (cp4) [context_patch, right=0.5cm of cp3] {};
    \node (cp5) [context_patch, right=0.5cm of cp4] {};
    \node (cp6) [context_patch, right=0.5cm of cp5] {};
    \node (cp7) [context_patch, right=0.5cm of cp6] {};
    \node (cp8) [context_patch, right=0.5cm of cp7] {};
    \node (cp9) [context_patch, right=0.5cm of cp8] {};
    
    \node (qp1) [context_patch, above=0cm of pps, xshift=4.3cm, yshift=-2cm] {};
    \node (qp2) [context_patch, right=0.5cm of qp1] {};
    \node (qp3) [context_patch, right=1.1cm of qp2] {};
    
    \node (pp1) [projection_patch, above=0.2cm of pps, xshift=-7cm] {};
    \node (ppd1) [projection_patch_description, left=0cm of pp1] {1};
    
    \node (pp2) [projection_patch, right=0.8cm of pp1] {};
    \node (ppd2) [projection_patch_description, left=0cm of pp2] {2};
    
    \node (pp3) [projection_patch, right=0.8cm of pp2] {};
    \node (ppd3) [projection_patch_description, left=0cm of pp3] {3};
    
    \node (pp4) [projection_patch, right=0.8cm of pp3] {};
    \node (ppd4) [projection_patch_description, left=0cm of pp4] {4};
    
    \node (pp5) [projection_patch, right=0.8cm of pp4] {};
    \node (ppd5) [projection_patch_description, left=0cm of pp5] {5};
    
    \node (pp6) [projection_patch, right=0.8cm of pp5] {};
    \node (ppd6) [projection_patch_description, left=0cm of pp6] {6};
    
    \node (pp7) [projection_patch, right=0.8cm of pp6] {};
    \node (ppd7) [projection_patch_description, left=0cm of pp7] {7};
    
    \node (pp8) [projection_patch, right=0.8cm of pp7] {};
    \node (ppd8) [projection_patch_description, left=0cm of pp8] {8};
    
    \node (pp9) [projection_patch, right=0.8cm of pp8] {};
    \node (ppd9) [projection_patch_description, left=0cm of pp9] {9};
    
    \node (q1) [projection_patch, above=0.2cm of pps, xshift=4.45cm] {};
    \node (qm1) [projection_patch_description, left=0cm of q1] {0};
    
    \node (q2) [projection_patch, right=0.8cm of q1] {};
    \node (qm2) [projection_patch_description, left=0cm of q2] {0};
    
    \node (q3) [projection_patch, right=1.4cm of q2] {};
    \node (qm3) [projection_patch_description, left=0cm of q3] {0};
    
    \node (transformer) [transformer, above=1cm of pps] {Transformer};
    
    \node (eb1) [embedding, above=0.2cm of transformer, xshift=4.45cm, fill=red!50] {};
    \node (eb2) [embedding, right=0.8cm of eb1, fill=green!50] {};
    \node (eb3) [embedding, right=1.4cm of eb2, fill=cyan!50] {};
    
    \node (readout) [readout, above=1.2cm of transformer] {PEs};
    \node (mlp1) [mlp, above=2.5cm of transformer, xshift=-5.9cm, fill=red!25] {PE1};
    \node (mlp2) [mlp, right=0.5cm of mlp1, fill=teal!25] {PE2};
    \node (mlp3) [mlp, right=0.5cm of mlp2, fill=blue!25] {PE3};
    \node (mlp4) [mlp, right=0.5cm of mlp3, fill=brown!25] {PE4};
    
    \node (output1) [output, above=1.5cm of mlp1, fill=red!50] {Left};
    \node (output2) [output, above=1.5cm of mlp2, fill=teal!50] {Right};
    \node (output3) [output, above=1.5cm of mlp3, fill=blue!50] {...};
    \node (output4) [output, above=1.5cm of mlp4, fill=brown!50] {...};
    
    \path [arrow] (cp1.north) -- (cp1|-pps.south);
    \path [arrow] (cp2.north) -- (cp2|-pps.south);
    \path [arrow] (cp3.north) -- (cp3|-pps.south);
    \path [arrow] (cp4.north) -- (cp4|-pps.south);
    \path [arrow] (cp5.north) -- (cp5|-pps.south);
    \path [arrow] (cp6.north) -- (cp6|-pps.south);
    \path [arrow] (cp7.north) -- (cp7|-pps.south);
    \path [arrow] (cp8.north) -- (cp8|-pps.south);
    \path [arrow] (cp9.north) -- (cp9|-pps.south);
    
    \path [arrow] (qp1.north) -- (qp1|-pps.south);
    \path [arrow] (qp2.north) -- (qp2|-pps.south);
    \path [arrow] (qp3.north) -- (qp3|-pps.south);
    
    \path [line] (pps.north-|pp1) -- (pp1.south);
    \path [line] (pps.north-|pp2) -- (pp2.south);
    \path [line] (pps.north-|pp3) -- (pp3.south);
    \path [line] (pps.north-|pp4) -- (pp4.south);
    \path [line] (pps.north-|pp5) -- (pp5.south);
    \path [line] (pps.north-|pp6) -- (pp6.south);
    \path [line] (pps.north-|pp7) -- (pp7.south);
    \path [line] (pps.north-|pp8) -- (pp8.south);
    \path [line] (pps.north-|pp9) -- (pp9.south);
    
    \path [line] (pps.north-|q1) -- (q1.south);
    \path [line] (pps.north-|q2) -- (q2.south);
    \path [line] (pps.north-|q3) -- (q3.south);
    
    \path [arrow] (pp1.north) -- (pp1|-transformer.south);
    \path [arrow] (pp2.north) -- (pp2|-transformer.south);
    \path [arrow] (pp3.north) -- (pp3|-transformer.south);
    \path [arrow] (pp4.north) -- (pp4|-transformer.south);
    \path [arrow] (pp5.north) -- (pp5|-transformer.south);
    \path [arrow] (pp6.north) -- (pp6|-transformer.south);
    \path [arrow] (pp7.north) -- (pp7|-transformer.south);
    \path [arrow] (pp8.north) -- (pp8|-transformer.south);
    \path [arrow] (pp9.north) -- (pp9|-transformer.south);
    
    \path [arrow] (q1.north) -- (q1|-transformer.south);
    \path [arrow] (q2.north) -- (q2|-transformer.south);
    \path [arrow] (q3.north) -- (q3|-transformer.south);
    
    \path [line] (transformer.north-|eb1) -- (eb1.south);
    \path [line] (transformer.north-|eb2) -- (eb2.south);
    \path [line] (transformer.north-|eb3) -- (eb3.south);
    
    \path [arrow, line width=0.4mm, color=red!50] (eb1.north) -- ++(0,0.4) -| ([yshift=0cm,xshift=-1cm]mlp1.south);
    \path [arrow, line width=0.4mm, color=red!50] (eb1.north) -- ++(0,0.4) -| ([yshift=0cm,xshift=-1cm]mlp2.south);
    \path [arrow, line width=0.4mm, color=red!50] (eb1.north) -- ++(0,0.4) -| ([yshift=0cm,xshift=-1cm]mlp3.south);
    \path [arrow, line width=0.4mm, color=red!50] (eb1.north) -- ++(0,0.4) -| ([yshift=0cm,xshift=-1cm]mlp4.south);
    
    \path [arrow, line width=0.4mm, color=teal] (eb2.north) -- ++(0,0.7) -| ([yshift=0cm,xshift=0cm]mlp1.south);
    \path [arrow, line width=0.4mm, color=teal] (eb2.north) -- ++(0,0.7) -| ([yshift=0cm,xshift=0cm]mlp2.south);
    \path [arrow, line width=0.4mm, color=teal] (eb2.north) -- ++(0,0.7) -| ([yshift=0cm,xshift=0cm]mlp3.south);
    \path [arrow, line width=0.4mm, color=teal] (eb2.north) -- ++(0,0.7) -| ([yshift=0cm,xshift=0cm]mlp4.south);
    
    \path [arrow, line width=0.4mm, color=cyan] (eb3.north) -- ++(0,1.0) -| ([yshift=0cm,xshift=1cm]mlp1.south);
    \path [arrow, line width=0.4mm, color=cyan] (eb3.north) -- ++(0,1.0) -| ([yshift=0cm,xshift=1cm]mlp2.south);
    \path [arrow, line width=0.4mm, color=cyan] (eb3.north) -- ++(0,1.0) -| ([yshift=0cm,xshift=1cm]mlp3.south);
    \path [arrow, line width=0.4mm, color=cyan] (eb3.north) -- ++(0,1.0) -| ([yshift=0cm,xshift=1cm]mlp4.south);
    
    \path [arrow, line width=0.2mm] (mlp1.north) -- ([yshift=0cm,xshift=0cm]output1.south);
    \path [arrow, line width=0.2mm] (mlp2.north) -- ([yshift=0cm,xshift=0cm]output2.south);
    \path [arrow, line width=0.2mm] (mlp3.north) -- ([yshift=0cm,xshift=0cm]output3.south);
    \path [arrow, line width=0.2mm] (mlp4.north) -- ([yshift=0cm,xshift=0cm]output4.south);

\end{tikzpicture}
}
\caption{\footnotesize The model structure. The input to the model consists of two major parts: (i) Context Patches: the cropped regions of the scene image, and (ii) Canonical Object Views: the descriptive image of the object to be queried. In the middle is the default Transformer model. Finally, there are four independent PEs, which correspond to different Predicates.
}
\label{fig:arch_detailed}
\end{figure}
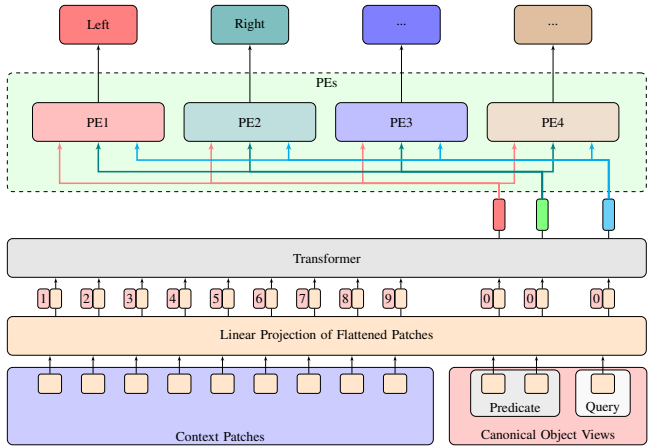

\section{VISUAL DEEP LEARNING MODELS AS PLANNING HEURISTICS}
Here, we describe the deep learning models that we have developed. 
These models accept images as input and output information that assist TAMP to improve planning time. 
\subsection{Embedding Model}
\subsubsection{The Vision Transformer (ViT)}
The inputs to the model are cropped image patches. 
These patches are of two types: the first type is the scene image and the second one is the canonical views of the objects that are queried (Fig.~\ref{fig:2}). 
The patches are first converted into an embedding, e.g., by a convolutional network. 
Then, the corresponding positional information is concatenated, which is later adopted in the transformer module. 

In the default ViT architecture~\cite{dosovitskiy2020image}, the embedding at position zero is critical. 
It is not derived from the input patch, but instead can be learned and then used for the classification.
This part is where our model differs from the original ViT.
Our implementation extends the use of location labels to enable the model to perform the query task in a given situation, i.e., to determine whether the query object needs to be involved in the process of completing the task given the target state.

The location information and the embedding are then passed into the transformer. 
The transformer here is identical to the original transformer structure used for natural language processing~\cite{vaswani2017attention}. 
However, unlike the traditional transformer, the vision transformer discards all the final embeddings of the patches from the scene image, and uses the final embedding of the zero position (Fig.~\ref{fig:arch_detailed}). 
A multilayer perceptron is finally implemented to use this embedding to predict the relevancy of the queried object for several predicates - we call this component the Predicate Extractor (PE).


\subsubsection{The Residual Neural Network (ResNet)}
\begin{figure*}
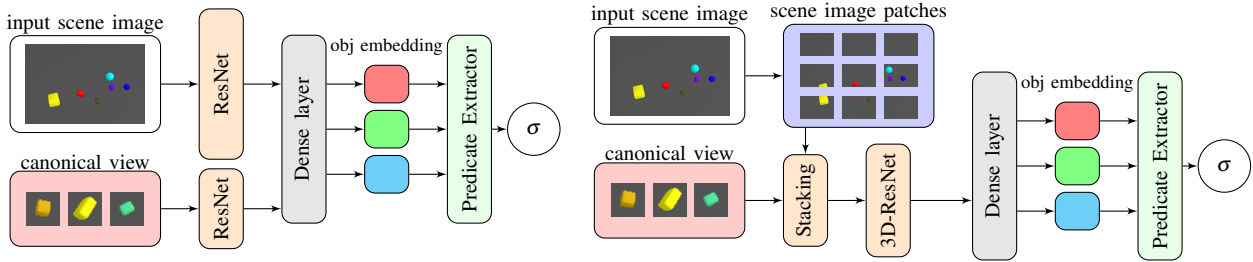

\centering
\begin{minipage}[c]{.38\linewidth}
\centering
\begin{tikzpicture}[node distance = 0cm, auto]
\node (scene_img) [input] {\includegraphics[width=0.9\linewidth]{img/scene_img.png}};
\node (scene_img_caption) [above=-0.1cm of scene_img] {\footnotesize input scene image};

\node (canonical_view) [input, below=0.5cm of scene_img, fill=red!20] 
    {\includegraphics[width=0.25\linewidth]{img/canonical_obj9.png}
     \includegraphics[width=0.25\linewidth]{img/canonical_obj3.png}
     \includegraphics[width=0.25\linewidth]{img/canonical_obj8.png}};
\node (canonical_view_caption) [above=-0.1cm of canonical_view] {\footnotesize canonical view};

\node (resnetl) [resnetl, right=0.5cm of scene_img, fill=orange!20] {\rotatebox{90}{\footnotesize ResNet}};
\node (resnets) [resnets, right=0.5cm of canonical_view, fill=orange!20] {\rotatebox{90}{\footnotesize ResNet}};
\node (transformer) [cnn, right=0.5cm of resnetl, yshift=-0.6cm, fill=gray!20] {\rotatebox{90}{\footnotesize Dense layer}};
\node (t1) [t, right=0.5cm of transformer, yshift=0.6cm, fill=red!50] {\rotatebox{90}{}};
\node (t2) [t, right=0.5cm of transformer, yshift=0cm, fill=green!50] {\rotatebox{90}{}};
\node (t3) [t, right=0.5cm of transformer, yshift=-0.6cm, fill=cyan!50] {\rotatebox{90}{}};
\node (embedding_caption) [above=0.01cm of t1] {\scriptsize obj embedding};
\node (readout) [cnn, right=0.5cm of t2, yshift=0cm, fill=green!10] {\rotatebox{90}{\footnotesize Predicate Extractor}};
\node (output_sigma) [output_sigma, right=0.2cm of readout] {\footnotesize $\sigma$};

\path [arrow] (scene_img.east) -- ([yshift=0cm,xshift=0cm]resnetl.west);
\path [arrow] (canonical_view.east) -- ([yshift=0cm,xshift=0cm]resnets.west);
\path [arrow] (resnetl.east) -- (resnetl-|transformer.west);
\path [arrow] (resnets.east) -- (resnets-|transformer.west);
\path [arrow] ([yshift=0.6cm,xshift=0cm]transformer.east) -- (t1.west);
\path [arrow] ([yshift=0cm,xshift=0cm]transformer.east) -- (t2.west);
\path [arrow] ([yshift=-0.6cm,xshift=0cm]transformer.east) -- (t3.west);

\path [arrow] (t1.east) -- ([yshift=0.6cm,xshift=0cm]readout.west);
\path [arrow] (t2.east) -- ([yshift=0cm,xshift=0cm]readout.west);
\path [arrow] (t3.east) -- ([yshift=-0.6cm,xshift=0cm]readout.west);

\path [arrow] (readout.east) -- (output_sigma.west);

\end{tikzpicture}
\end{minipage}%
\ 
\begin{minipage}[c]{.6\linewidth}
\centering
\begin{tikzpicture}[node distance = 0cm, auto]
\node (scene_img) [input] {\includegraphics[width=0.9\linewidth]{img/scene_img.png}};
\node (scene_img_caption) [above=-0.1cm of scene_img] {\footnotesize input scene image};

\node (canonical_view) [input, below=0.5cm of scene_img, fill=red!20] 
    {\includegraphics[width=0.25\linewidth]{img/canonical_obj9.png}
     \includegraphics[width=0.25\linewidth]{img/canonical_obj3.png}
     \includegraphics[width=0.25\linewidth]{img/canonical_obj8.png}};
\node (canonical_view_caption) [above=-0.1cm of canonical_view] {\footnotesize canonical view};

\node (scene_img_patches) [input, right=0.5cm of scene_img, fill=blue!20] { 
    \adjincludegraphics[width=.25\linewidth,trim={{.0\width} {.66\height} {.66\width} {.0\height}},clip]{img/scene_img.png}
    \adjincludegraphics[width=.25\linewidth,trim={{.33\width} {.66\height} {.33\width} {.0\height}},clip]{img/scene_img.png}
    \adjincludegraphics[width=.25\linewidth,trim={{.66\width} {.66\height} {.0\width} {.0\height}},clip]{img/scene_img.png}
    
    \adjincludegraphics[width=.25\linewidth,trim={{.0\width} {.33\height} {.66\width} {.33\height}},clip]{img/scene_img.png}
    \adjincludegraphics[width=.25\linewidth,trim={{.33\width} {.33\height} {.33\width} {.33\height}},clip]{img/scene_img.png}
    \adjincludegraphics[width=.25\linewidth,trim={{.66\width} {.33\height} {.0\width} {.33\height}},clip]{img/scene_img.png}
    
    \adjincludegraphics[width=.25\linewidth,trim={{.0\width} {.0\height} {.66\width} {.66\height}},clip]{img/scene_img.png}
    \adjincludegraphics[width=.25\linewidth,trim={{.33\width} {.0\height} {.33\width} {.66\height}},clip]{img/scene_img.png}
    \adjincludegraphics[width=.25\linewidth,trim={{.66\width} {.0\height} {.0\width} {.66\height}},clip]{img/scene_img.png}
    };
\node (scene_img_patches_caption) [above=-0.1cm of scene_img_patches] {\footnotesize scene image patches};
\node (stacking) [resnets, right=0.5cm of canonical_view, fill=orange!20] {\rotatebox{90}{\footnotesize Stacking}};

\node (resnet) [resnets, right=0.5cm of stacking, fill=orange!20] {\rotatebox{90}{\footnotesize 3D-ResNet}};
\node (transformer) [cnn, right=0.5cm of scene_img_patches, yshift=-1.2cm, fill=gray!20] {\rotatebox{90}{\footnotesize Dense layer}};
\node (t1) [t, right=0.5cm of transformer, yshift=0.6cm, fill=red!50] {\rotatebox{90}{}};
\node (t2) [t, right=0.5cm of transformer, yshift=-0cm, fill=green!50] {\rotatebox{90}{}};
\node (t3) [t, right=0.5cm of transformer, yshift=-0.6cm, fill=cyan!50] {\rotatebox{90}{}};
\node (embedding_caption) [above=0.01cm of t1] {\scriptsize obj embedding};
\node (readout) [cnn, right=0.5cm of t2, yshift=0cm, fill=green!10] {\rotatebox{90}{\footnotesize Predicate Extractor}};
\node (output_sigma) [output_sigma, right=0.2cm of readout] {\footnotesize $\sigma$};

\path [arrow] (scene_img.east) -- ([yshift=0cm,xshift=0cm]scene_img_patches.west);
\path [arrow] (canonical_view.east) -- ([yshift=0cm,xshift=0cm]stacking.west);
\path [arrow] (scene_img_patches.south -| stacking.north) -- (stacking.north);
\path [arrow] (stacking.east) -- (resnet.west);
\path [arrow] (resnet.east) -- (resnet-|transformer.west);
\path [arrow] (transformer.east|-t1.west) -- (t1.west);
\path [arrow] (transformer.east|-t2.west) -- (t2.west);
\path [arrow] (transformer.east|-t3.west) -- (t3.west);

\path [arrow] (t1.east) -- (t1-|readout.west);
\path [arrow] (t2.east) -- (t2-|readout.west);
\path [arrow] (t3.east) -- (t3-|readout.west);

\path [arrow] (readout.east) -- (output_sigma.west);

\end{tikzpicture}
\end{minipage}
\caption{Proposed architectures based on ResNet: (left) a model based on a conventional ResNet, and (right) a model based on a 3D-ResNet. 
}
\label{fig:resnet_arch}
\end{figure*}

ResNets extract features by convolutional and residual blocks~\cite{he2016deep}.
We built two versions (Fig.~\ref{fig:resnet_arch}):
For the first one ((D)efault-ResNet), two different sizes of ResNets are used to analyze the scene image and the canonical views separately.
A fully linked network then produces three embeddings similar to the ViT-based model. 
The PE eventually receives the generated embedding data to extract the relevance information for the query object given a task described by canonical object views.
In the second one, which we named three-dimensional ResNet, we first segment the scene image into patches as well, and stack them with the canonical view. 
Then, a residual neural network based on a 3D convolutional algorithm is used to extract the features (Fig.~\ref{fig:resnet_arch}).

The features output from the residual-based neural network model can also be considered as embedded information and utilized by the \textit{Predicate Extractor (PE)}. 

\subsection{Embedding Information Extraction}
One option is to model \textit{PE} to have the same output dimensionality as the number of the predicates that we are concerned with: for example, the predicates \textit{on-the-left}, \textit{on-the-right}, \textit{in-front-of} and \textit{behind}. 
In our current architecture, there is a single \textit{PE} for each predicate as shown in Fig.~\ref{fig:arch_detailed}.

We trained the model with batch polling method.
This method is based on a polling approach (Fig.~\ref{fig:training_method_2}), where the data corresponding to different PEs are taken out in batches and used for the corresponding training, thus updating the corresponding PEs and the transformer they share. 
With the polling method, the transformer is updated with data from multiple predicates simultaneously at each gradient step, leading to a more reliable optimization.

\subsection{Details of the Models} \label{sec:model_details}
The goal of this study is to learn a classifier $t$ predicting an object set ${O} \subset \mathcal{O}$ that comprises the objects manipulated by the actions $a_{1:K}$ generated by a feasible nonlinear program $P(g, S)$.
In order to achieve this objective, the key consideration is how to represent the scene $S$ and the desired configuration $\mathcal{S}_{goal}(g)$ as inputs to the classifier. This representation should not only facilitate the training of a precise classifier but also allow it to perform well on different scenes, including those with varying numbers of objects.
Due to the fixed-size input characteristics of the image sensor, we can argue that the image sensor meets our requirements.
Formally, we assume that there is a generative process to produce an image $I \in \mathbb{R}^{w \times h \times 3}$ from the scene $S$, either via a rendering engine in simulation or a camera in the real world. 
We choose an image acquired by a tilted (virtual) camera to capture as much geometric information as possible.

Thanks to the attention mechanism, all we need to identify for our query is a canonical view per object $q \in \mathbb{R}^{32 \times 32}$ obtained in advance. 
Note that the dimensions of the canonical views here are same as the patches of scene image, which are also 32 by 32.

The input to the classifier $t(I, p_{0,...,L}, q_i)$ is the image $I$ as three channel image $I \in \mathbb{R}^{w \times h \times 3}$ and canonical views for the objects, which, together with the goal predicate, describe the goal and the queried object, $p_{0,...,L} \in \mathbb{R}^{32 \times 32}$, $q_i$, respectively ($L=1$ for our PEs in this study). 
Fig.~\ref{fig:1} visualizes these input images, and Fig.~\ref{fig:2} describes our architecture with a high level of abstraction. 
The network is trained to output the probability that the queried objects will be included in the manipulation plan which is the result of the nonlinear program $P(g, S)$
\begin{equation} \label{eq:4}
    t(I, p_{0,...,L}, q_{i})=p(Z_{S}(i)|F_{S}(a_{1:K})=1)_{o \in \mathcal{O}}
\end{equation}
using the standard weighted binary cross-entropy loss
\begin{equation} \label{eq:5}
\begin{aligned}
    L(w) = \sum_{i \in \mathcal{O}} & \eta Z_{S}(i)\log(t(I, p_{0,...,L}, q_{i};w)) \\
                                    & + (1 - Z_{S}(i))\log(1 - t(I, p_{0,...L}, q_{i};w)).
\end{aligned}
\end{equation}

The training data $\mathcal{D}={(I_{i}, \mathrm{P}_{i}, Q_{i})}_{i=1}^{d}$ consists of scene images $I_{i}$ with the canonical views of the predicate related objects $\mathrm{P}_{i}$ and query object $Q_{i}$. 
In our experiments the majority of the objects are irrelevant to the goal state. Therefore, adjusting the value of weighting factor $\eta \leq 1$ in eq.~\ref{eq:5} becomes important to balance the loss, otherwise the classifier could indiscriminately predict the biased label to achieve high accuracies.
If the prediction $t(I, p_{0,...L}, q_{i})$ for object $i$ is lower than the threshold $\beta \in \mathbb{R}$, then we decide that $Z_{S}(i) = 1$, i.e., the object is relevant to the action sequence.
In the experiments, we discuss the influence of this threshold.

For the ViT and the 3D-ResNet models, the scene image is first cropped into small patches of a predefined size, we use 32 by 32 in our experiments. 
For the ViT-based model, the patches of the scene images, along with the canonical views associated with the predicates and the query objects, are then fed into the first layer of the CNN network, which is used to transform them into an embedding with a size of 768 required by the transformer.
The embedding converted from the scene image will then be attached with the corresponding position label, while the embedding generated from the canonical views will only be attached with position label zero. 
In essence, we only use the embeddings that capture the correlation between context patches and canonical views, and also between canonical views themselves, while discarding the correlation between context patches (for details, see~\cite{dosovitskiy2020image, yuan2022sornet}). 
These embeddings with labels attached are then fed into the standard transformer. 
Finally, we take out the output corresponding to the canonical views and feed them into the final fully connected layer, which has two layers and 512 neurons per layer (which we call the \textit{Predicate Extractor}).

\section{GUIDING MOTION PLANNING USING LEARNED
VISUAL HEURISTICS}
The classifier defined in \eqref{eq:4} can be used not only to predict the relevance of a goal state depicted by a goal predicate within a scene, but also to serve as a heuristic for optimizing the mixed-integer program presented in \eqref{eq:1}.
This is because it produces a score that can be normalized and interpreted as the probability of relevance.
The goal of this heuristic is to find a set of objects $O \subset \mathcal{O}(S)$ such that the NLP $P(g, S)$ is feasible with all actions $a_{1:K}$ of the objects that are operated on $o \in O$ (i.e., $\theta(a) \in O$). 
In this way, the search space for finding a feasible NLP $P(g, S)$ could be reduced to a certain level, which improves the time complexity of the whole computation.

\subsection{Admissible Heuristic} \label{sec:ad_heuristic}
The classifier may serve as an admissible heuristic. 
In this context, the term \textit{admissible} suggests that errors occurring during classification cannot inhibit the discovery of a solution to a viable problem, denoted as $P(g, S)$.
This is realized by first evaluating $t(I, p_{0,...,L}, q_{i}) = p_{i}$ for all $o_{i} \in \mathcal{O}$, where $q_{i}$ is the canonical view of $o_{i}$. 
The NLP $P(g, S)$ first searches for a feasible path with actions only considering the objects in $O \subset \mathcal{O}$. 
If it is feasible, then a solution will be given, otherwise the whole object set will be used for searching for a feasible path. 
In the instance of an infeasible problem, where no viable path leads to the goal state, all action-object tuples will be examined. Consequently, employing the learned network as an admissible heuristic offers no advantage in such scenarios.

\subsection{Non-Admissible Heuristic}\label{sec:nonad_heuristic}
Similar to real-world settings, many objects are irrelevant for the goal state in our experiments. 
In addition, as long as the object set predicted by the classifier does not contain all of the objects that the NLP has to consider in order to find a feasible path, the NLP will clearly be unable to find a feasible path. 
In Sec.~\ref{sec:model_details} we mentioned the hyperparameter $\beta$ as the threshold for the binary output $Z_{S}(i)$ (Eq.~\ref{eq:4}).
In practice we can reduce $\beta$ to a certain amount to increase the size of object set, and to have a higher probability to fully cover the object set which is required by a feasible path finding procedure.
Because the predicted object set is still a subset of the whole object set $\mathcal{O}$, the search space remains limited, thus the time complexity is expected to be improved. 
During the experiments, we assess and analyze both the false irrelevant rate and the time savings concerning the relevance threshold. 
For $\beta = 0$, this method becomes identical to the admissible heuristic of Section~\ref{sec:ad_heuristic}.

\begin{figure} 
\centering
\begin{minipage}[c]{0.32\linewidth}
\centering
\subfloat[Overview]{\includegraphics[width=\linewidth]{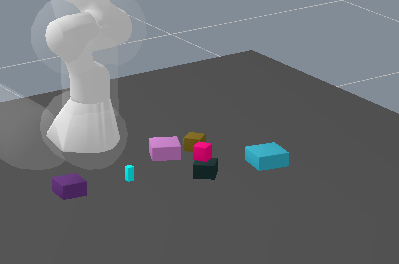}}
\end{minipage}%
\ 
\begin{minipage}[c]{0.32\linewidth}
\centering
\subfloat[Interference]{\includegraphics[width=\linewidth]{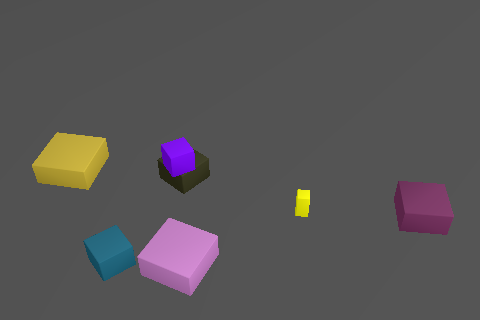}}
\end{minipage}%
\
\begin{minipage}[c]{0.32\linewidth}
\centering
\subfloat[Non-interference]{\includegraphics[width=\linewidth]{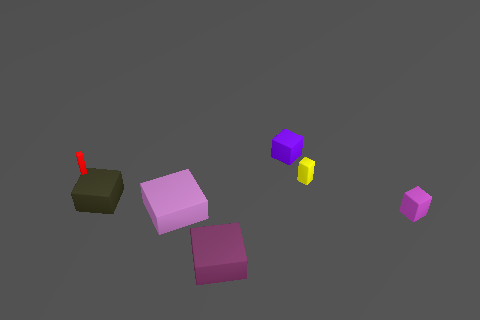}}
\end{minipage}%
\caption{An example of scene for the \textit{on top} predicate.}
\label{fig:above}
\end{figure}

\begin{table*}
\centering
\begin{tabular}{ccccccccc}
\hline
 &
\multicolumn{2}{c}{} &\multicolumn{3}{c}{accuracy of classifier} & \multicolumn{3}{c}{performance on LGP} \\
\multirow{2}{*}{\makecell{Test \\ data}} & predicate & $\eta$ & \multirow{2}{*}{true rel.} & \multirow{2}{*}{true irrel.} & \multirow{2}{*}{total accuracy} &  \multirow{2}{*}{runtime(s)} & \multirow{2}{*}{our runtime(s)} & \multirow{2}{*}{speed-up}\\
\multicolumn{9}{c}{}\\
\hline
\multirow{6}{*}{same-plane}  
& on-the-left   & 0.86 & 87\% & 94\% & 93\% & $3.06 \pm 0.100$ & $1.83 \pm 0.065$ & 40.2\%\\  
& on-the-right  & 0.86 & 88\% & 94\% & 93\% & $3.37 \pm 0.104$ & $1.87 \pm 0.061$ & 44.8\%\\ 
& in-front      & 0.86 & 92\% & 96\% & 95\% & $3.64 \pm 0.105$ & $2.09 \pm 0.062$ & 42.6\%\\ 
& behind        & 0.86 & 92\% & 95\% & 95\% & $3.36 \pm 0.104$ & $1.92 \pm 0.065$ & 42.8\%\\  \cline{2-9}
& total  & 0.86 & 90\% & 95\% & 94\% & $4.73$ & $1.93$ & 42.6\%\\ 
\hline
different-plane
& on & 0.66 & 76\% & 89\% & 87\% & $4.47 \pm 0.103$ & $3.70 \pm 0.101$ & 17.2\% \\ 
\hline  
\end{tabular}
\caption{Accuracies on all validation sets and Running time comparison of LGP using vision transformer based deep learning model and original LGP. We report 67\% confidence intervals.}
\label{tbl:accuracies}
\end{table*}

\section{Experiments}

In our experiments, we consider two types of predicates: predicates describing objects on the same horizontal plane, such as \textit{on-the-left} and \textit{on-the-right}, and predictates describing objects in different planes, here we choose \textit{on top}.

\subsection{Data Generation}
For all experiments, we place objects randomly on the table and a camera looking down 45 degrees to get the scene image.
We assume that each distinct object has a different color.
For the \textit{on-top} predicate, we also randomly select a target object and then randomly place another object on top of the target object. 
This is done in order to increase the likelihood that an object in the scene is task-relevant.
Randomly distributing all objects would in almost all cases result in all objects being on the same horizontal plane, which would mean that for the \textit{on-top} predicate, no other objects apart from those being part of the goal are relevant.
Placing one object on top of the target ensures that the classifier is forced to make a decision on whether it needs to be removed.
This will depend on whether there is enough space for the goal object or not, and is therefore a nontrivial visual task. 
Fig. \ref{fig:1} depicts a typical scene for the same-plane case, where the task is to place object $c$ relative to object $d$ (e.g., on the left-hand side of object $d$). The classifier gets an extra input object $e$, and outputs the probability that object $e$ has to be interacted with in order to complete the mentioned task.
All tasks in such configurations would take between 2 and 6 steps to complete, with the number of steps required depending on the size of the area occupied by the target location, the size of the objects to be placed, and the spatial arrangement.
Fig~\ref{fig:above} shows an example of the different-plane case.

For generating labels, we rely on the outcome of LGP. 
Given this initial scene and a goal predicate indicating a spatial relationship between two objects, the LGP agent has to figure out both the sequence of high-level decisions and the motion plans that can realize those tasks jointly.
If the agent successfully finds a solution, the objects which have been manipulated according to the motion plan will be labeled as relevant and recorded.
If no plan has been found, nothing is recorded.

As we mentioned before, for any given goal predicate, the majority of objects in the same scene $S$ are irrelevant.
Training a single model to output all $4$ predicates simultaneously would mean that there are $2^4=16$ different labels, some of which (for example the \textit{all true} case) never occur, resulting in a very imbalanced dataset.
We therefore decide to train 4 different model heads, the PEs, on a single binary label that is independently balanced.
We still share weights of the model body, the details of which are described in section ~\ref{sec:training_modes}.
This also means that the model scales gracefully to a large number of predicates.

\begin{figure}[t]
\centering
\resizebox{1\linewidth}{!}{
\begin{tikzpicture}[node distance = 0cm, auto]
    \node (transformer) [transformer, above=1cm of pps] {Transformer};
    
    \node (readout) [readout, above=1.2cm of transformer] {PEs};
    \node (mlp1) [mlp, above=2.5cm of transformer, xshift=-5.9cm, fill=red!25] {PE1};
    \node (mlp2) [mlp, right=0.5cm of mlp1, fill=teal!25] {PE2};
    \node (mlp3) [mlp, right=0.5cm of mlp2, fill=blue!25] {PE3};
    \node (mlp4) [mlp, right=0.5cm of mlp3, fill=brown!25] {PE4};
    
    \node (dataset, minimum height=2.6cm) [dataset, above=1.2cm of readout] {Datasets};
    \node (data1) [mlp, above=2.5cm of readout, xshift=-5.9cm, fill=red!50] {dataset1};
    \node (data2) [mlp, right=0.5cm of data1, fill=teal!50] {dataset2};
    \node (data3) [mlp, right=0.5cm of data2, fill=blue!50] {dataset3};
    \node (data4) [mlp, right=0.5cm of data3, fill=brown!50] {dataset4};
    

    \path [arrow, line width=0.1mm, color=red] ([yshift=0cm,xshift=-1cm]mlp1.south) -- ++(0,-1.6) -| ([yshift=0cm,xshift=-1cm]transformer.north);
    \path [arrow, line width=0.1mm, color=teal] ([yshift=0cm,xshift=-1cm]mlp2.south) -- ++(0,-0.7) -| ([yshift=0cm,xshift=-1cm]transformer.north);
    \path [arrow, line width=0.1mm, color=blue] ([yshift=0cm,xshift=-1cm]mlp3.south) -- ++(0,-0.9) -| ([yshift=0cm,xshift=-1cm]transformer.north);
    \path [arrow, line width=0.1mm, color=brown] ([yshift=0cm,xshift=-1cm]mlp4.south) -- ++(0,-1.1) -| ([yshift=0cm,xshift=-1cm]transformer.north);
    
    \path [arrow, line width=0.1mm, color=red] ([yshift=0cm,xshift=0cm]mlp1.south) -- ++(0,-1.6) -| ([yshift=0cm,xshift=0cm]transformer.north);
    \path [arrow, line width=0.1mm, color=teal] ([yshift=0cm,xshift=0cm]mlp2.south) -- ++(0,-0.7) -| ([yshift=0cm,xshift=0cm]transformer.north);
    \path [arrow, line width=0.1mm, color=blue] ([yshift=0cm,xshift=0cm]mlp3.south) -- ++(0,-0.9) -| ([yshift=0cm,xshift=0cm]transformer.north);
    \path [arrow, line width=0.1mm, color=brown] ([yshift=0cm,xshift=0cm]mlp4.south) -- ++(0,-1.1) -| ([yshift=0cm,xshift=0cm]transformer.north);
    
    \path [arrow, line width=0.1mm, color=red] ([yshift=0cm,xshift=1cm]mlp1.south) -- ++(0,-1.6) -| ([yshift=0cm,xshift=1cm]transformer.north);
    \path [arrow, line width=0.1mm, color=teal] ([yshift=0cm,xshift=1cm]mlp2.south) -- ++(0,-0.7) -| ([yshift=0cm,xshift=1cm]transformer.north);
    \path [arrow, line width=0.1mm, color=blue] ([yshift=0cm,xshift=1cm]mlp3.south) -- ++(0,-0.9) -| ([yshift=0cm,xshift=1cm]transformer.north);
    \path [arrow, line width=0.1mm, color=brown] ([yshift=0cm,xshift=1cm]mlp4.south) -- ++(0,-1.1) -| ([yshift=0cm,xshift=1cm]transformer.north);

    \path [arrow, line width=0.1mm, color=red] ([yshift=0cm,xshift=-1cm]data1.south) -- ++(0,-1.6) -| node [arrow_txt] {1.} ([yshift=0cm,xshift=-1cm]mlp1.north);
    \path [arrow, line width=0.1mm, color=teal] ([yshift=0cm,xshift=-1cm]data2.south) -- ++(0,-1.6) -| node [arrow_txt] {2.} ([yshift=0cm,xshift=-1cm]mlp2.north);
    \path [arrow, line width=0.1mm, color=blue] ([yshift=0cm,xshift=-1cm]data3.south) -- ++(0,-1.6) -| node [arrow_txt] {3.} ([yshift=0cm,xshift=-1cm]mlp3.north);
    \path [arrow, line width=0.1mm, color=brown] ([yshift=0cm,xshift=-1cm]data4.south) -- ++(0,-1.6) -| node [arrow_txt] {4.} ([yshift=0cm,xshift=-1cm]mlp4.north);
    
    \path [arrow, line width=0.1mm, color=red] ([yshift=0cm,xshift=0cm]data1.south) -- ++(0,-1.6) -| node [arrow_txt] {5.} ([yshift=0cm,xshift=0cm]mlp1.north);
    \path [arrow, line width=0.1mm, color=teal] ([yshift=0cm,xshift=0cm]data2.south) -- ++(0,-1.6) -| node [arrow_txt] {6.} ([yshift=0cm,xshift=0cm]mlp2.north);
    \path [arrow, line width=0.1mm, color=blue] ([yshift=0cm,xshift=0cm]data3.south) -- ++(0,-1.6) -| node [arrow_txt] {7.} ([yshift=0cm,xshift=0cm]mlp3.north);
    \path [arrow, line width=0.1mm, color=brown] ([yshift=0cm,xshift=0cm]data4.south) -- ++(0,-1.6) -| node [arrow_txt] {8.} ([yshift=0cm,xshift=0cm]mlp4.north);
    
    \path [arrow, line width=0.1mm, color=red] ([yshift=0cm,xshift=1cm]data1.south) -- ++(0,-1.6) -| node [arrow_txt] {9...} ([yshift=0cm,xshift=1cm]mlp1.north);
    \path [arrow, line width=0.1mm, color=teal] ([yshift=0cm,xshift=1cm]data2.south) -- ++(0,-1.6) -| node [arrow_txt] {10...} ([yshift=0cm,xshift=1cm]mlp2.north);
    \path [arrow, line width=0.1mm, color=blue] ([yshift=0cm,xshift=1cm]data3.south) -- ++(0,-1.6) -| node [arrow_txt] {11...} ([yshift=0cm,xshift=1cm]mlp3.north);
    \path [arrow, line width=0.1mm, color=brown] ([yshift=0cm,xshift=1cm]data4.south) -- ++(0,-1.6) -| node [arrow_txt] {12...} ([yshift=0cm,xshift=1cm]mlp4.north);
    
\end{tikzpicture}
}
\caption{The batch polling training Procedure}
\label{fig:training_method_2}
\end{figure}
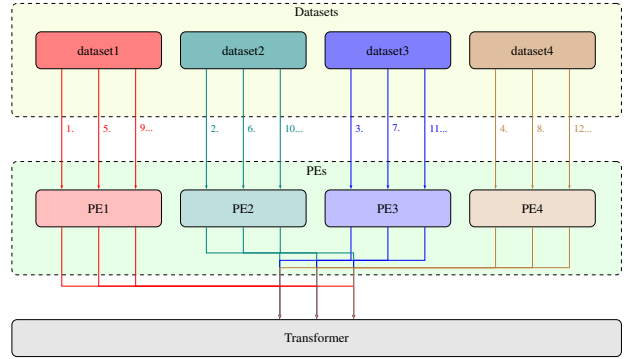
\subsection{Training the multi-head model}
\label{sec:training_modes}

We use $4$ different model heads, one for each predicate.
The body of the model is shared between these.
The model is trained end-to-end by periodically taking a batch of data from each predicate's corresponding training set (Fig.~\ref{fig:training_method_2}).
In an alternative approach, we also tried training each predicate head to convergence, before moving on to the next one.
Unsurprisingly, the first approach we described works significantly better and used throughout the paper.



\begin{table}
\centering
\begin{tabular}{c c c c c} 
\hline
& predicate & relevant (pct.) & irrelevant (pct.) \\ [0.5ex] 
\hline
\multirow{4}{*}{Training set} 
& on-the-left     & 1235677 (13.7\%) & 7764323 (86.3\%) \\ 
& on-the-right    & 1239533 (13.8\%) & 7760467 (86.2\%) \\
& in-front        & 1239388 (13.8\%) & 7760612 (86.2\%) \\
& behind          & 1248629 (13.9\%) & 7751371 (86.1\%) \\
\cline{2-4}
\multirow{4}{*}{Test set} 
& on-the-left     & 12448 (13.8\%) & 77552 (86.2\%) \\
& on-the-right    & 12557 (14.0\%) & 77443 (86.0\%) \\
& in-front        & 12401 (13.8\%) & 77599 (86.2\%) \\
& behind          & 12495 (13.9\%) & 77505 (86.1\%) \\
\hline
Training set & on   & 3650855 (30.4\%) & 8349145 (69.6\%) \\ 
\cline{2-4}
Test set     & on   & 34689 (28.9\%) & 85311 (71.1\%) \\
\hline
\end{tabular}
\caption{Distribution of labels in each dataset}
\label{tbl:datasets_overview}
\end{table}

\subsection{Predicates in the same plane}
For the first phase of the experiments, we consider datasets for \textit{on-the-left}, \textit{on-the-right}, \textit{in-front} and \textit{behind}.
Each dataset consists of 80,000 scenes in which there are between 3 and 10 objects in the scenes. Both of the training set and test set are generated with the same strategy.
As mentioned before, the data imbalance is quite severe, with only about 14\% of the queries being relevant and about 86\% of the queries being irrelevant.
Their data distribution is shown in the Table~\ref{tbl:datasets_overview}.
We use $\eta = 0.86$ (Eq.~\ref{eq:5}) to counteract the imbalance in the training dataset.
The network is trained with the \textbf{Adam} optimizer, using a learning rate of 0.00001 and a batch size of 50.

\subsection{\textit{On-top} predicate}

The training data contains 160,000 scenes, and the scenes also have various number of objects.
30\% of the objects were relevant, 70\% were irrelevant. 
The same imbalanced data issue arises here as well, and we use $\eta = 0.66$. 
Again, the network is trained with the \textbf{Adam} optimizer, using a learning rate of 0.00001 and a batch size of 50.

\begin{figure}
\centering
\begin{minipage}[t]{.48\linewidth}
\centering
\begin{tikzpicture}[trim axis left, trim axis right]
\pgfplotsset{
    scale only axis,
    xmin=-0.05, xmax=0.7,
    legend style={font=\footnotesize, nodes={scale=0.8, transform shape}, fill=none},
    y axis style/.style={
        yticklabel style=#1,
        ylabel style=#1,
        y axis line style=#1,
        ytick style=#1
   },
   every x tick label/.append style={font=\scriptsize,},
   every y tick label/.append style={font=\scriptsize,},
}
\begin{axis}[
    title={},
    tick label style={font=\scriptsize, /pgf/number format/fixed},
    axis y line*=left,
    x label style={at={(axis description cs:0.5,0.1)},anchor=north},
    y label style={at={(axis description cs:0.2,.5)},anchor=south},
    xlabel={\scriptsize Threshold $\beta$},
    width= 0.8\linewidth,
    ymin=0, ymax=1.05,
    xtick={0, 0.1, 0.2, 0.3, 0.4, 0.5, 0.6, 0.7},
    ytick={0, 0.2, 0.4, 0.6, 0.8, 1},
]
\addplot[smooth,color=DeepSkyBlue3,]
    coordinates {
    (0,   0.8)
    (0.065, 0.81)
    (0.13,  0.82)
    (0.195, 0.83)
    (0.26,  0.84)
    (0.325, 0.85)
    (0.39,  0.86)
    (0.455, 0.87)
    (0.52,  0.88)
    (0.585, 0.89)
    (0.65,  0.90)
    };\label{trplot}
\addplot[smooth,color=Firebrick3,]
    coordinates {
    (0,   0.91)
    (0.065, 0.90)
    (0.13,  0.89)
    (0.195, 0.88)
    (0.26,  0.86)
    (0.325, 0.85)
    (0.39,  0.84)
    (0.455, 0.82)
    (0.52,  0.81)
    (0.585, 0.79)
    (0.65,  0.77)
    };\label{tiplot}
\addplot[smooth, color=black,]
    coordinates {
    (0,   0.88)
    (0.065, 0.87)
    (0.13,  0.87)
    (0.195, 0.86)
    (0.26,  0.86)
    (0.325, 0.85)
    (0.39,  0.84)
    (0.455, 0.84)
    (0.52,  0.83)
    (0.585, 0.82)
    (0.65,  0.81)
    };\label{total_acc}

\end{axis}

\begin{axis}[
    tick label style={font=\scriptsize},
    axis y line*=right,
    axis x line=none,
    width= 0.8\linewidth,
    ymin=0, ymax=1.05,
    ytick={0, 0.2, 0.4, 0.6, 0.8, 1},
    yticklabels={,,},
    legend style={at={(0.25,0.69)}, anchor=north west},
    every y tick label/.append style={font=\scriptsize, xshift=-0.5ex},
    scaled ticks=false, 
    tick label style={/pgf/number format/fixed}
]

\addlegendimage{/pgfplots/refstyle=trplot}\addlegendentry{true relevant}
\addlegendimage{/pgfplots/refstyle=tiplot}\addlegendentry{true irrelevant}
\addlegendimage{/pgfplots/refstyle=total_acc}\addlegendentry{total accuracy}
\addplot[smooth,Sienna1] 
  coordinates{
    (0, 0.2)
    (0.065, 0.19)
    (0.13, 0.17)
    (0.195, 0.17)
    (0.26, 0.16)
    (0.325, 0.15)
    (0.39, 0.14)
    (0.455, 0.13)
    (0.52, 0.12)
    (0.585, 0.11)
    (0.65, 0.10)
}; \addlegendentry{false irrelevant}

\end{axis}
\end{tikzpicture}
\caption{Classifier accuracies}
\label{fig:accuracies_graph}
\end{minipage}
\ 
\begin{minipage}[t]{.48\linewidth}
\centering
\begin{tikzpicture}[trim axis left, trim axis right]
\pgfplotsset{
    scale only axis,
    xmin=-0.05, xmax=0.7,
    legend style={font=\footnotesize, nodes={scale=0.8, transform shape}},
    y axis style/.style={
        yticklabel style=#1,
        ylabel style=#1,
        y axis line style=#1,
        ytick style=#1
   },
    every x tick label/.append style={font=\scriptsize,},
    every y tick label/.append style={font=\scriptsize,},
}
\begin{axis}[
    title={},
    tick label style={font=\scriptsize, /pgf/number format/fixed},
    axis y line*=left,
    ylabel near ticks, yticklabel pos=right,
    x label style={at={(axis description cs:0.5,0.1)},anchor=north},
    y label style={at={(axis description cs:1.3,.5)},anchor=south},
    xlabel={\scriptsize Threshold $\beta$},
    yticklabel={$\pgfmathprintnumber{\tick}\%$},
    width= 0.8\linewidth,
    ymin=-5, ymax=40,
    xtick={0, 0.1, 0.2, 0.3, 0.4, 0.5, 0.6, 0.7},
    ytick={0, 10, 20, 30, 40},
]

\end{axis}

\begin{axis}[
    tick label style={font=\scriptsize, /pgf/number format/fixed},
    axis y line*=right,
    axis x line=none,
    width= 0.8\linewidth,
    ymin=-5, ymax=40,
    yticklabels={,,},
    ytick={0, 10, 20, 30, 40},
    every y tick label/.append style={font=\scriptsize, xshift=-0.5ex},
]

\addplot[smooth, color=Red1,]
    coordinates {
    (0.0, 17.2)
    (0.065, 19.0)
    (0.13,  31.0)
    (0.195, 16.0)
    (0.26,  23.3)
    (0.325, 22.1)
    (0.39,  23.1)
    (0.455, 22.3)
    (0.52,  22.1)
    (0.585, 18.2)
    (0.65,  20.4)
    };\addlegendentry{speedup}

\end{axis}
\end{tikzpicture}
\caption{LGP speed-up}
\label{fig:lgp_spped_up}
\end{minipage}
\end{figure}

\subsection{Efficiency Improvement on LGP}
Finally, we compare the improvement brought by the model for sequential planning of actions using LGP.
We found out that the accuracies for different models are comparable (Tbl.~\ref{tbl:accuracies_diff_model}). 
Since the runtime of LGP is only dependent on the accuracy of the model, we will focus on the results obtained using the model based on the Transformer.
The results for both same-plane and different-plane datasets are shown in Tbl.~\ref{tbl:accuracies}.
We use the unmodified LGP as the baseline in our experiments.
We do not show the impact of false irrelevant predictions of the classifier on the performance of the LGP.
The influence of false irrelevant predictions on the runtime depends on other design choices like maximum runtime settings, which we factor out here.

The model performs better on the first dataset than on the second dataset, where we find an over 40\% improvement on LGP. 
The second dataset has only 76 percent of true relevant due to the complexity of the predicates and the occlusion problem of the visual sensors, but it still provides a 17\% improvement.

As mentioned in section ~\ref{sec:nonad_heuristic}, $\beta$ directly affects the percentage of false irrelevant predictions, and the presence of false irrelevant prediction means that LGP has a high probability of not finding a feasible manipulation sequence.

We analyse how $\beta$ would affect the runtime of LGP, and the results are shown in Fig.~\ref{fig:accuracies_graph} and Fig.~\ref{fig:lgp_spped_up}. 
The data in Fig.~\ref{fig:accuracies_graph} is obtained by using the classifier trained on the data for the $on-top$ predicate. 
As $\beta$ increases, the model will adopt a more conservative strategy, so that more data will be marked as relevant.
This also leads to a decrease in the number of false irrelevant cases. 
Since the action planning of LGP only considers the objects marked as "relevant", missing the objects that should be considered will result in LGP not finding a feasible plan. 
By adjusting $\beta$, we can avoid this situation to some extent.

\subsection{Comparison of base models}
In the last part of the experiments we compare the training results based on different visual deep learning models. 
All results in the experiments are obtained for the more complicated $on-top$ predicate dataset.
\begin{table}[!ht]
\centering
\begin{tabular}{ccccc}
\hline
\multicolumn{2}{c}{} &\multicolumn{3}{c}{accuracy of classifier} \\
\multirow{2}{*}{\makecell{Base Model}} & $\eta$ & \multirow{2}{*}{true rel.} & \multirow{2}{*}{true irrel.} & \multirow{2}{*}{total acc.} \\
\multicolumn{5}{c}{}\\
\hline
Transformer & 0.66 & 81\% & 89\% & 87\%  \\ 
3D-ResNet & 0.66 & 79\% & 90\% & 87\%  \\ 
Default ResNet & 0.66 & 81\% & 91\% & 88\%  \\ 
\hline   
\end{tabular}
\caption{Comparison of accuracy rates using different base models}
\label{tbl:accuracies_diff_model}
\end{table}

The best results are obtained using Default ResNet as the base model.
This is followed by the 3D-ResNet-based model, and finally the Transformer-based model.
It is worth noting that even though we used very different vision models, the final results are not significantly different.

\section{Conclusion}
As other visual models, our model can not circumvent problems due to occlusion and other ambiguities. Regardless of how the vision sensors are placed, it is possible for smaller objects to be obscured by larger objects, as is often the case for the \textit{on-top} predicate.
Furthermore, our dataset is relatively homogeneous in terms of object color selection, and the images are not affected by ambient lighting.
While this can likely be addressed in a straightforward way by using photorealistic renderings and by randomizing lights and colors, it currently limits the number of scenarios the model can be used in.

Sequential manipulation planning becomes inefficient when applied to scenes where many non-relevant objects are present.
We address this in this study by using a learned visual heuristic to classify objects into non-relevant and relevant objects, given a symbolic goal.
We use a deep learning network that uses the scene image and canonical views describing the goal and a queried object as input, and makes predictions about the involvement of this object in the scene. 
These predictions can be directly used in the process of sequential manipulation planning, thus directly improving the efficiency of planning.


We experimented on two different types of predicates: the first for spatial relations on the same plane and the second for spatial relations on different planes. 
They both perform well in terms of prediction accuracy and reduce the time required for sequential manipulation planning.


\bibliographystyle{IEEEtranN}
\footnotesize
\bibliography{master_thesis}

\begin{thebibliography}{32}
\providecommand{\natexlab}[1]{#1}
\providecommand{\url}[1]{#1}
\csname url@samestyle\endcsname
\providecommand{\newblock}{\relax}
\providecommand{\bibinfo}[2]{#2}
\providecommand{\BIBentrySTDinterwordspacing}{\spaceskip=0pt\relax}
\providecommand{\BIBentryALTinterwordstretchfactor}{4}
\providecommand{\BIBentryALTinterwordspacing}{\spaceskip=\fontdimen2\font plus
\BIBentryALTinterwordstretchfactor\fontdimen3\font minus
  \fontdimen4\font\relax}
\providecommand{\BIBforeignlanguage}[2]{{%
\expandafter\ifx\csname l@#1\endcsname\relax
\typeout{** WARNING: IEEEtranN.bst: No hyphenation pattern has been}%
\typeout{** loaded for the language `#1'. Using the pattern for}%
\typeout{** the default language instead.}%
\else
\language=\csname l@#1\endcsname
\fi
#2}}
\providecommand{\BIBdecl}{\relax}
\BIBdecl

\bibitem[Wells et~al.(2019)Wells, Dantam, Shrivastava, and
  Kavraki]{wells2019learning}
A.~M. Wells, N.~T. Dantam, A.~Shrivastava, and L.~E. Kavraki, ``Learning
  feasibility for task and motion planning in tabletop environments,''
  \emph{IEEE robotics and automation letters}, vol.~4, no.~2, pp. 1255--1262,
  2019.

\bibitem[Toussaint et~al.(2018)Toussaint, Allen, Smith, and
  Tenenbaum]{toussaint2018differentiable}
M.~A. Toussaint, K.~R. Allen, K.~A. Smith, and J.~B. Tenenbaum,
  ``Differentiable physics and stable modes for tool-use and manipulation
  planning,'' 2018.

\bibitem[Rodr{\i}guez et~al.(2019)Rodr{\i}guez, Nottensteiner, Leidner,
  Ka{\ss}ecker, Stulp, and Albu-Sch{\"a}ffer]{rodriguez2019iteratively}
I.~Rodr{\i}guez, K.~Nottensteiner, D.~Leidner, M.~Ka{\ss}ecker, F.~Stulp, and
  A.~Albu-Sch{\"a}ffer, ``Iteratively refined feasibility checks in robotic
  assembly sequence planning,'' \emph{IEEE Robotics and Automation Letters},
  vol.~4, no.~2, pp. 1416--1423, 2019.

\bibitem[Driess et~al.(2019)Driess, Oguz, and
  Toussaint]{driess2019hierarchical}
D.~Driess, O.~Oguz, and M.~Toussaint, ``Hierarchical task and motion planning
  using logic-geometric programming (hlgp),'' in \emph{RSS Workshop on Robust
  Task and Motion Planning}, 2019.

\bibitem[Kase et~al.(2020)Kase, Paxton, Mazhar, Ogata, and
  Fox]{kase2020transferable}
K.~Kase, C.~Paxton, H.~Mazhar, T.~Ogata, and D.~Fox, ``Transferable task
  execution from pixels through deep planning domain learning,'' in \emph{2020
  IEEE International Conference on Robotics and Automation (ICRA)}.\hskip 1em
  plus 0.5em minus 0.4em\relax IEEE, 2020, pp. 10\,459--10\,465.

\bibitem[Driess et~al.(2021)Driess, Ha, Tedrake, and
  Toussaint]{driess2021learning}
D.~Driess, J.-S. Ha, R.~Tedrake, and M.~Toussaint, ``Learning geometric
  reasoning and control for long-horizon tasks from visual input,'' in
  \emph{2021 IEEE International Conference on Robotics and Automation
  (ICRA)}.\hskip 1em plus 0.5em minus 0.4em\relax IEEE, 2021, pp.
  14\,298--14\,305.

\bibitem[Driess et~al.(2020)Driess, Oguz, Ha, and Toussaint]{driess2020deep}
D.~Driess, O.~Oguz, J.-S. Ha, and M.~Toussaint, ``Deep visual heuristics:
  Learning feasibility of mixed-integer programs for manipulation planning,''
  in \emph{2020 IEEE International Conference on Robotics and Automation
  (ICRA)}.\hskip 1em plus 0.5em minus 0.4em\relax IEEE, 2020, pp. 9563--9569.

\bibitem[Yuan et~al.(2022)Yuan, Paxton, Desingh, and Fox]{yuan2022sornet}
W.~Yuan, C.~Paxton, K.~Desingh, and D.~Fox, ``Sornet: Spatial object-centric
  representations for sequential manipulation,'' in \emph{Conference on Robot
  Learning}.\hskip 1em plus 0.5em minus 0.4em\relax PMLR, 2022, pp. 148--157.

\bibitem[Alili et~al.(2010)Alili, Pandey, Sisbot, and
  Alami]{alili2010interleaving}
S.~Alili, A.~K. Pandey, E.~A. Sisbot, and R.~Alami, ``Interleaving symbolic and
  geometric reasoning for a robotic assistant,'' in \emph{ICAPS Workshop on
  Combining Action and Motion Planning}, 2010.

\bibitem[Srivastava et~al.(2014)Srivastava, Fang, Riano, Chitnis, Russell, and
  Abbeel]{srivastava2014combined}
S.~Srivastava, E.~Fang, L.~Riano, R.~Chitnis, S.~Russell, and P.~Abbeel,
  ``Combined task and motion planning through an extensible planner-independent
  interface layer,'' in \emph{2014 IEEE international conference on robotics
  and automation (ICRA)}.\hskip 1em plus 0.5em minus 0.4em\relax IEEE, 2014,
  pp. 639--646.

\bibitem[de~Silva et~al.(2013)de~Silva, Pandey, Gharbi, and
  Alami]{de2013towards}
L.~de~Silva, A.~K. Pandey, M.~Gharbi, and R.~Alami, ``Towards combining htn
  planning and geometric task planning,'' \emph{arXiv preprint
  arXiv:1307.1482}, 2013.

\bibitem[Dantam et~al.(2018)Dantam, Kingston, Chaudhuri, and
  Kavraki]{dantam2018incremental}
N.~T. Dantam, Z.~K. Kingston, S.~Chaudhuri, and L.~E. Kavraki, ``An incremental
  constraint-based framework for task and motion planning,'' \emph{The
  International Journal of Robotics Research}, vol.~37, no.~10, pp. 1134--1151,
  2018.

\bibitem[Sim{\'e}on et~al.(2004)Sim{\'e}on, Laumond, Cort{\'e}s, and
  Sahbani]{simeon2004manipulation}
T.~Sim{\'e}on, J.-P. Laumond, J.~Cort{\'e}s, and A.~Sahbani, ``Manipulation
  planning with probabilistic roadmaps,'' \emph{The International Journal of
  Robotics Research}, vol.~23, no. 7-8, pp. 729--746, 2004.

\bibitem[Erdem et~al.(2011)Erdem, Haspalamutgil, Palaz, Patoglu, and
  Uras]{erdem2011combining}
E.~Erdem, K.~Haspalamutgil, C.~Palaz, V.~Patoglu, and T.~Uras, ``Combining
  high-level causal reasoning with low-level geometric reasoning and motion
  planning for robotic manipulation,'' in \emph{2011 IEEE International
  Conference on Robotics and Automation}.\hskip 1em plus 0.5em minus
  0.4em\relax IEEE, 2011, pp. 4575--4581.

\bibitem[Lagriffoul et~al.(2012)Lagriffoul, Dimitrov, Saffiotti, and
  Karlsson]{lagriffoul2012constraint}
F.~Lagriffoul, D.~Dimitrov, A.~Saffiotti, and L.~Karlsson, ``Constraint
  propagation on interval bounds for dealing with geometric backtracking,'' in
  \emph{2012 IEEE/RSJ International Conference on Intelligent Robots and
  Systems}.\hskip 1em plus 0.5em minus 0.4em\relax IEEE, 2012, pp. 957--964.

\bibitem[Lagriffoul et~al.(2014)Lagriffoul, Dimitrov, Bidot, Saffiotti, and
  Karlsson]{lagriffoul2014efficiently}
F.~Lagriffoul, D.~Dimitrov, J.~Bidot, A.~Saffiotti, and L.~Karlsson,
  ``Efficiently combining task and motion planning using geometric
  constraints,'' \emph{The International Journal of Robotics Research},
  vol.~33, no.~14, pp. 1726--1747, 2014.

\bibitem[Lozano-P{\'e}rez and Kaelbling(2014)]{lozano2014constraint}
T.~Lozano-P{\'e}rez and L.~P. Kaelbling, ``A constraint-based method for
  solving sequential manipulation planning problems,'' in \emph{2014 IEEE/RSJ
  International Conference on Intelligent Robots and Systems}.\hskip 1em plus
  0.5em minus 0.4em\relax IEEE, 2014, pp. 3684--3691.

\bibitem[Shoukry et~al.(2016)Shoukry, Nuzzo, Saha, Sangiovanni-Vincentelli,
  Seshia, Pappas, and Tabuada]{shoukry2016scalable}
Y.~Shoukry, P.~Nuzzo, I.~Saha, A.~L. Sangiovanni-Vincentelli, S.~A. Seshia,
  G.~J. Pappas, and P.~Tabuada, ``Scalable lazy smt-based motion planning,'' in
  \emph{2016 IEEE 55th Conference on Decision and Control (CDC)}.\hskip 1em
  plus 0.5em minus 0.4em\relax IEEE, 2016, pp. 6683--6688.

\bibitem[Hadfield-Menell et~al.(2016)Hadfield-Menell, Lin, Chitnis, Russell,
  and Abbeel]{hadfield2016sequential}
D.~Hadfield-Menell, C.~Lin, R.~Chitnis, S.~Russell, and P.~Abbeel, ``Sequential
  quadratic programming for task plan optimization,'' in \emph{2016 IEEE/RSJ
  International Conference on Intelligent Robots and Systems (IROS)}.\hskip 1em
  plus 0.5em minus 0.4em\relax IEEE, 2016, pp. 5040--5047.

\bibitem[Toussaint(2015)]{toussaint2015logic}
M.~Toussaint, ``Logic-geometric programming: An optimization-based approach to
  combined task and motion planning,'' in \emph{Twenty-Fourth International
  Joint Conference on Artificial Intelligence}, 2015.

\bibitem[Toussaint et~al.(2020)Toussaint, Ha, and
  Driess]{toussaint2020describing}
M.~Toussaint, J.-S. Ha, and D.~Driess, ``Describing physics for physical
  reasoning: Force-based sequential manipulation planning,'' \emph{IEEE
  Robotics and Automation Letters}, vol.~5, no.~4, pp. 6209--6216, 2020.

\bibitem[Ha et~al.(2020)Ha, Driess, and Toussaint]{ha2020probabilistic}
J.-S. Ha, D.~Driess, and M.~Toussaint, ``A probabilistic framework for
  constrained manipulations and task and motion planning under uncertainty,''
  in \emph{2020 IEEE International Conference on Robotics and Automation
  (ICRA)}.\hskip 1em plus 0.5em minus 0.4em\relax IEEE, 2020, pp. 6745--6751.

\bibitem[Hogan et~al.(2018)Hogan, Grau, and Rodriguez]{hogan2018reactive}
F.~R. Hogan, E.~R. Grau, and A.~Rodriguez, ``Reactive planar manipulation with
  convex hybrid mpc,'' in \emph{2018 IEEE International Conference on Robotics
  and Automation (ICRA)}.\hskip 1em plus 0.5em minus 0.4em\relax IEEE, 2018,
  pp. 247--253.

\bibitem[Deits and Tedrake(2014)]{deits2014footstep}
R.~Deits and R.~Tedrake, ``Footstep planning on uneven terrain with
  mixed-integer convex optimization,'' in \emph{2014 IEEE-RAS international
  conference on humanoid robots}.\hskip 1em plus 0.5em minus 0.4em\relax IEEE,
  2014, pp. 279--286.

\bibitem[Vaswani et~al.(2017)Vaswani, Shazeer, Parmar, Uszkoreit, Jones, Gomez,
  Kaiser, and Polosukhin]{vaswani2017attention}
A.~Vaswani, N.~Shazeer, N.~Parmar, J.~Uszkoreit, L.~Jones, A.~N. Gomez,
  {\L}.~Kaiser, and I.~Polosukhin, ``Attention is all you need,''
  \emph{Advances in neural information processing systems}, vol.~30, 2017.

\bibitem[Dosovitskiy et~al.(2020)Dosovitskiy, Beyer, Kolesnikov, Weissenborn,
  Zhai, Unterthiner, Dehghani, Minderer, Heigold, Gelly,
  et~al.]{dosovitskiy2020image}
A.~Dosovitskiy, L.~Beyer, A.~Kolesnikov, D.~Weissenborn, X.~Zhai,
  T.~Unterthiner, M.~Dehghani, M.~Minderer, G.~Heigold, S.~Gelly \emph{et~al.},
  ``An image is worth 16x16 words: Transformers for image recognition at
  scale,'' \emph{arXiv preprint arXiv:2010.11929}, 2020.

\bibitem[Carion et~al.(2020)Carion, Massa, Synnaeve, Usunier, Kirillov, and
  Zagoruyko]{carion2020end}
N.~Carion, F.~Massa, G.~Synnaeve, N.~Usunier, A.~Kirillov, and S.~Zagoruyko,
  ``End-to-end object detection with transformers,'' in \emph{European
  conference on computer vision}.\hskip 1em plus 0.5em minus 0.4em\relax
  Springer, 2020, pp. 213--229.

\bibitem[Guo et~al.(2022)Guo, Yang, Chen, Ma, Xie, and Pu]{Guo20222ndPS}
Y.~Guo, S.~Yang, W.~Chen, L.~Ma, D.~Xie, and S.~Pu, ``2nd place solution for
  iccv 2021 vipriors image classification challenge: An attract-and-repulse
  learning approach,'' \emph{ArXiv}, vol. abs/2206.06168, 2022.

\bibitem[Nguyen et~al.(2020)Nguyen, Oguz, Hartmann, and
  Toussaint]{nguyen2020self}
S.~Nguyen, O.~S. Oguz, V.~N. Hartmann, and M.~Toussaint, ``Self-supervised
  learning of scene-graph representations for robotic sequential manipulation
  planning.'' in \emph{CoRL}, 2020, pp. 2104--2119.

\bibitem[He et~al.(2016)He, Zhang, Ren, and Sun]{he2016deep}
K.~He, X.~Zhang, S.~Ren, and J.~Sun, ``Deep residual learning for image
  recognition,'' in \emph{Proceedings of the IEEE conference on computer vision
  and pattern recognition}, 2016, pp. 770--778.

\bibitem[Toussaint and Lopes(2017)]{toussaint2017multi}
M.~Toussaint and M.~Lopes, ``Multi-bound tree search for logic-geometric
  programming in cooperative manipulation domains,'' in \emph{2017 IEEE
  International Conference on Robotics and Automation (ICRA)}.\hskip 1em plus
  0.5em minus 0.4em\relax IEEE, 2017, pp. 4044--4051.

\bibitem[Mason(1985)]{mason1985mechanics}
M.~Mason, ``The mechanics of manipulation,'' in \emph{Proceedings. 1985 IEEE
  International Conference on Robotics and Automation}, vol.~2.\hskip 1em plus
  0.5em minus 0.4em\relax IEEE, 1985, pp. 544--548.

\end{thebibliography}



\end{document}